\documentclass[10pt,twocolumn,letterpaper]{article}

\usepackage{wacv}
\usepackage{times}
\usepackage{epsfig}
\usepackage{graphicx}
\usepackage{amsmath}
\usepackage{amssymb}

\usepackage{graphicx}
\usepackage{comment}
\usepackage{amsmath,amssymb} 
\usepackage{color}
\usepackage{wrapfig}
\usepackage{booktabs}
\usepackage{cite}
\usepackage{enumitem}
\usepackage{caption}
\usepackage{pifont}
\newcommand{\xmark}{\ding{55}}%

\def\wacvPaperID{1088} 

\wacvfinalcopy 

\ifwacvfinal
\def\assignedStartPage{1} 
\fi

\ifwacvfinal
\usepackage[breaklinks=true,bookmarks=false]{hyperref}
\else
\usepackage[pagebackref=true,breaklinks=true,colorlinks,bookmarks=false]{hyperref}
\fi

\ifwacvfinal
\else
\fi

\begin{document}

\title{Hierarchically Decoupled Spatial-Temporal Contrast for Self-supervised Video Representation Learning}

\author{Zehua Zhang~~~~~~~David Crandall\\
Indiana University Bloomington\\
{\tt\small \{zehzhang, djcran\}@indiana.edu}
}

\maketitle

\begin{abstract}
We present a novel technique for self-supervised video representation learning by: (a) decoupling the learning objective into two contrastive subtasks respectively emphasizing spatial and temporal features, and (b) performing it hierarchically to encourage multi-scale understanding. Motivated by their effectiveness in supervised learning, we first introduce spatial-temporal feature learning decoupling and hierarchical learning to the context of unsupervised video learning. We show by experiments that augmentations can be manipulated as regularization to guide the network to learn desired semantics in contrastive learning, and we propose a way for the model to separately capture spatial and temporal features at multiple scales. We also introduce an approach to overcome the problem of divergent levels of instance invariance at different hierarchies by modeling the invariance as loss weights for objective re-weighting. Experiments on downstream action recognition benchmarks on UCF101 and HMDB51 show that our proposed Hierarchically Decoupled Spatial-Temporal Contrast (HDC) makes substantial improvements over directly learning spatial-temporal features as a whole and achieves competitive performance when compared with other state-of-the-art unsupervised methods. Code will be made available.
\end{abstract}

\section{Introduction}
\label{sec:intro}

\begin{figure}[t]
    \begin{center}
       \includegraphics[width=0.48\textwidth]{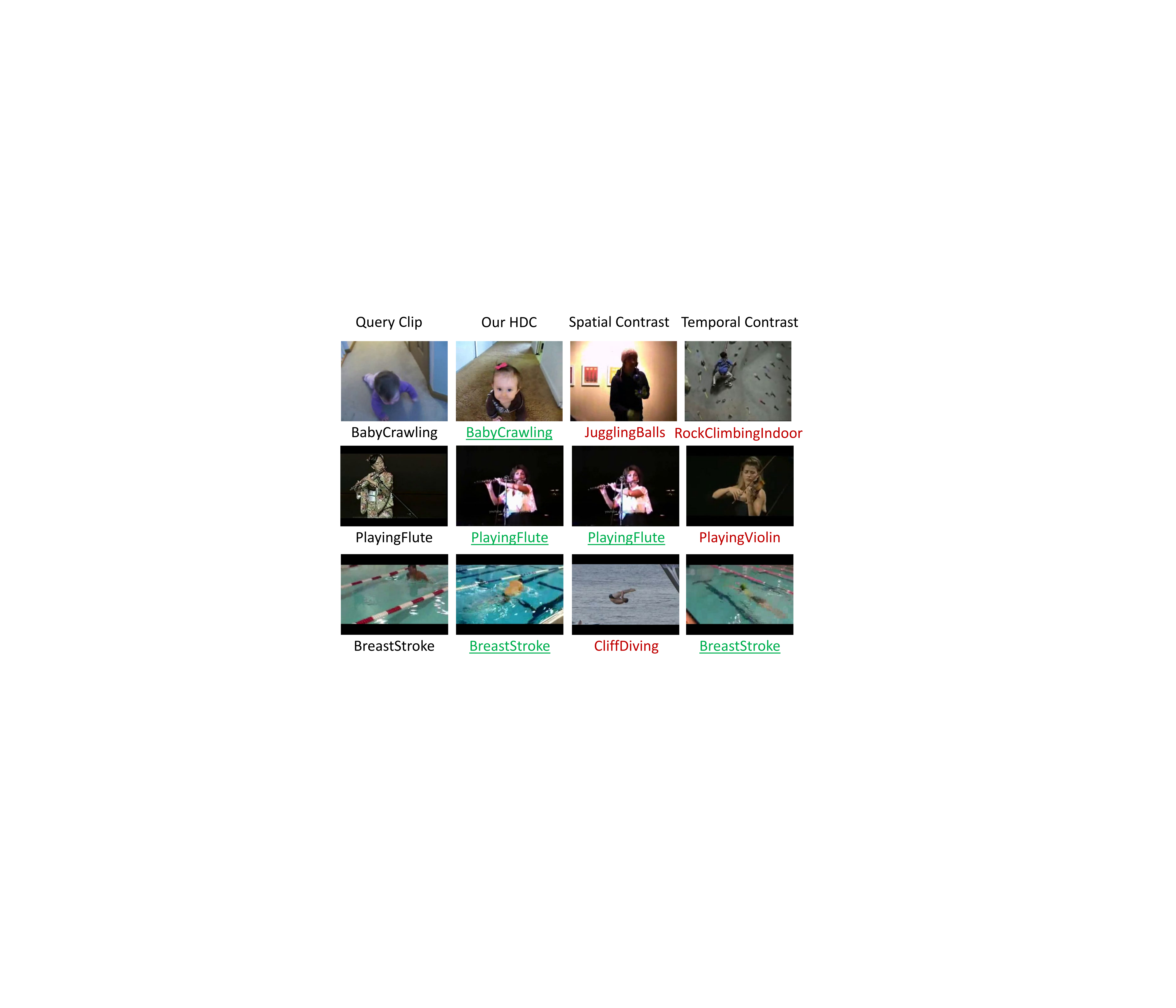}
    \end{center}
    \vspace{-16pt} 
    \caption{Sample results of nearest neighbor retrieval on UCF101 split 1 demonstrate the effectiveness of our  approach for decoupling spatial-temporal feature learning. We compare the top-1 retrieved clip from our full model (\textit{HDC}) with the clip retrieved by models performing only one subtask of HDC  (Spatial Contrast (\textit{SC}) or Temporal Contrast  (\textit{TC})). Below each clip is its corresponding action category, with green underlined text indicating a correct retrieval and red text indicating an incorrect retrieval. During retrieval, SC tends to focus more on spatial similarity with the query clip, while TC focuses  more on temporal similarity. For example, in the first row, SC's retrieval is perhaps based on spatial similarity of people indoors, while TC's retrieval may be due to temporal similarity of crawling. Since our proposed HDC solves SC and TC hierarchically and simultaneously, it is able to capture both spatial and temporal semantics and made the correct retrieval.}  
    \label{fig:pre} 
    \vspace{-12pt}
\end{figure}

As a solution to the growing need for large-scale labeled data for
training complex neural network
models~\cite{alexnet,resnet,inceptionv1,c3d,i3d}, unsupervised
representation learning aims to learn good feature embeddings from
data without annotations.  Using the learned representations as
initialization, downstream tasks only need to be fine-tuned on a
relatively small labeled dataset in order to yield reasonable
performance. Much recent progress in unsupervised image representation
learning~\cite{wu2018unsupervised,ye2019unsupervised,bachman2019learning,hjelm2018learning,oord2018representation,henaff2019data,he2019momentum,misra2019self,tian2019contrastive}
is driven by contrastive learning. While they solve the same pretext task of
instance-level variant matching, these methods differ in how they
obtain variant embeddings of the same instance, e.g., using
augmentations~\cite{wu2018unsupervised,ye2019unsupervised,bachman2019learning,hjelm2018learning},
future representations~\cite{oord2018representation}, or momentum
features~\cite{he2019momentum}. By optimizing a contrastive
loss~\cite{hadsell2006dimensionality}, they are essentially maximizing
intra-instance embedding similarity and minimizing inter-instance
embedding similarity, which leads to a spread-out feature
space~\cite{zhang2017learning}.

Compared with the success of contrastive learning for images, most
state-of-the-art unsupervised video representation learning work
relies on context-based proxy tasks such as time arrow
classification~\cite{timearrow} or clip order
prediction~\cite{xu2019self}. While the results of some recent
methods~\cite{sun2019learning, han2019video, cyclecontrast, Han20}
suggest the potential of contrastive learning for videos, they work
in a ``one for all'' manner by solving a single contrastive learning
task with the activation maps from the final layer, expecting the model to
capture all the features through the learning procedure. As a result,
the learned model may lack a general understanding of spatial and
temporal semantics, and instead just memorize spatial-temporal combinations.

In this paper, we argue that good video representations should be able
to capture spatial and temporal features in a more general form at
multiple scales, and thus it is  helpful to decompose the overall
goal of learning spatial-temporal features into hierarchical subtasks,
respectively emphasizing spatial and temporal features.
%
%
%
%
To this end, we present Hierarchically Decoupled Spatial-Temporal Contrast (HDC), 
in which we decouple the learning objective into separate
subtasks of Spatial Contrast and Temporal
Contrast and perform the learning hierarchically. 

Neural networks are notorious for learning shortcuts to
``cheat''~\cite{doersch2015unsupervised,timearrow}: if
there is an easy way to solve the problem, the network will hardly
try to find a more complex solution. We make use of this property for
the decoupling and direct the network to learn different features by
manipulating augmentations. In particular, in the Spatial Contrast
subtask, we deliberately provide a shortcut by creating augmented
variants with only spatial augmentations (random spatial cropping,
color jittering, etc.). As the timestamps of the query clip and its
augmented copy are the same, it is possible to solve the matching task
based merely on consistency of spatial semantics, and thus the network
will try to ``cheat'' by focusing more on spatial features. In
Temporal Contrast learning, we randomly select a new clip from the
video of the query clip (random temporal cropping) before applying
spatial augmentations in order to obtain a variant whose spatial
semantics are as different from the query clip as possible. Since
spatial context may vary dramatically after applying the temporal and
spatial transformations, the model is prevented from ``cheating''
through spatial similarity and encouraged to rely more on similarity
of temporal semantics to solve the pretext task. Fig.~\ref{fig:pre}
shows nearest neighbor retrieval results demonstrating the
effectiveness of our approach for the decoupling, and suggests the potential of directing the network to learn desired features through manipulating augmentations as regularization.

In order to capture multi-scale features, we further perform Spatial
Contrast and Temporal Contrast learning hierarchically by optimizing
towards a compound loss. During hierarchical learning, we model the
significance of instance-wise consistency in a given layer with
different weights, because features from different layers do not share
the same level of invariance against
augmentations~\cite{zeiler2014visualizing}.

\newenvironment{itemize*}%
               {\begin{itemize}[leftmargin=*]%
                   \setlength{\itemsep}{0pt}%
                   \setlength{\parskip}{0pt}}%
               {\end{itemize}}

In summary, our contributions are as follows:
\begin{itemize*}
 \item We demonstrate the effectiveness of spatial-temporal feature learning decoupling and hierarchical learning in the context of unsupervised learning for the first time.
 \item We show that the network can be guided to learn desired features in contrastive learning by manipulating augmentations as regularization, and introduce a new way for the network to separately capture spatial and temporal semantics in self-supervised video representation learning.
 \item We propose an approach to improve hierarchical contrastive learning by modeling the divergent levels of invariance in different layers as different loss weights.
 \item By optimizing a novel compound loss, our Hierarchically Decoupled Spatial-Temporal Contrast (HDC) outperforms
   other unsupervised methods and sets a new
   state-of-the-art on downstream tasks of action recognition on
   UCF101 and HMDB51.
\end{itemize*}

\section{Related Work}

\newcommand{\xhdr}[1]{\vspace{6pt}\noindent{\textbf{#1}}}
\xhdr{Unsupervised Video Representation Learning} 
was originally
based on input
reconstruction~\cite{vincent2008extracting,ranzato2007unsupervised,lee2007efficient,le2011learning,hinton2006reducing,hinton2006fast},
while more recent methods derive
implicit pseudo-labels from the
unlabeled data to use as self-supervision signals for the
corresponding pretext task~\cite{raina2007self,wang2015unsupervised}.
 For example, several models use
chronological order of video frames to define proxy tasks
such as frame order prediction~\cite{lee2017unsupervised} or
verification~\cite{misra2016shuffle,fernando2017self}, clip
order prediction~\cite{xu2019self}, and time arrow
classification~\cite{timearrow}. Other pretext tasks, such as
spatial-temporal jigsaw~\cite{kim2019self}, future
prediction~\cite{vondrick2016generating,srivastava2015unsupervised,han2019video,mathieu2015deep,vondrick2016anticipating,lotter2016deep},
temporal correspondence
estimation~\cite{wang2019learning,jayaraman2016slow,wang2015unsupervised,isola2015learning},
audio-video clustering~\cite{alwassel2019self}, video
colorization~\cite{vondrick2018tracking}, motion and appearance
statistics prediction~\cite{wang2019self}, and loss distillation across multiple tasks~\cite{piergiovanni2020evolving} have also been explored.



\xhdr{Contrastive Learning}~\cite{hadsell2006dimensionality} is very
effective for unsupervised representation learning for
images~\cite{wu2018unsupervised,ye2019unsupervised,bachman2019learning,hjelm2018learning,oord2018representation,henaff2019data,he2019momentum,misra2019self,tian2019contrastive}.
These methods try to learn a feature space in which variants of the
same sample are close together while variants from different samples
are far apart, and mainly differ in how they create the variants. For
example, Oord~\etal~\cite{oord2018representation} predict the future
in the latent space as a variant of the real embeddings of the future.
In the video domain, Han~\etal~\cite{han2019video} predict dense
feature maps of future clips and match them with corresponding real
embeddings from other distractions. This idea is further extended to a
memory-augmented version~\cite{Han20} for
improvement. Sun~\etal~\cite{sun2019learning} proposed to use
bidirectional transformers for multi-modal contrastive learning from
text and videos. A cycle-contrastive loss inspired by
CycleGAN~\cite{CycleGAN2017} is presented by
Kong~\etal~\cite{cyclecontrast} to use  the relationship
between videos and frames. Recasens \etal~\cite{recasens2021broaden} and Wang \etal~\cite{wang2021long} both explore a new pretext task of matching short-term view to long-term view of the same video. Tschannen \etal~\cite{tschannen2019self}
also use video-induced invariance to formulate a pretext task
based on contrastive learning.  Despite its motivation, interestingly,
it aims to learn image representations instead of video embeddings.
There is more discussion in Sec.~\ref{sec:diff}.



\section{Our Method}

To explore the potential of self-supervised video representation
learning merely from RGB clips, we formulate a novel pretext task of
Hierarchically Decoupled Spatial-Temporal Contrast (HDC) in which
we maximize intra-instance representation similarity and minimize
inter-instance representation similarity spatially (Fig.~\ref{fig:stc}), temporally (Fig.~\ref{fig:stc}), and
hierarchically (Fig.~\ref{fig:histc}).

\subsection{Decoupled Contrast}
\label{sec:stc}
\begin{figure*}[t]
    \begin{center}
       \includegraphics[width=0.85\textwidth]{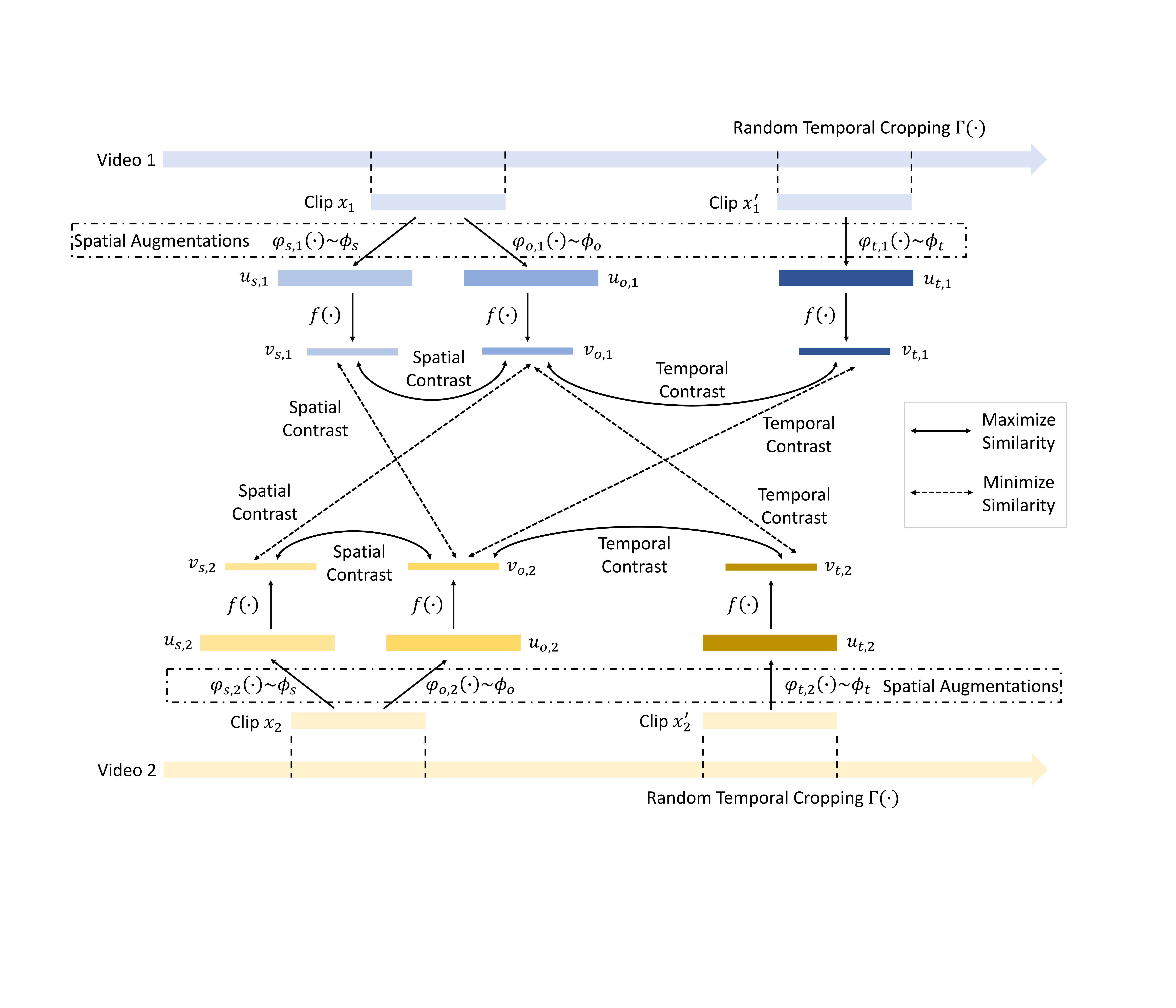}
    \end{center}
    \vspace{-12pt} 
    \caption{A simple example of how our Decoupled Contrast 
    works between two sample videos. In Spatial Contrast (SC), two sets of spatial transformations, $\varphi_{o,i}$ and $\varphi_{s,i}$, are sampled from families of candidate spatial transformations $\Phi_o$ and $\Phi_s$, and applied to the same clip $x_i$ to obtain variants $u_{o,i}$ and $u_{s,i}$ for SC learning. In Temporal Contrast (TC), we first extract another clip $x'_i$ from the same video as $x_i$ by random temporal cropping $\Gamma(\cdot)$, and then apply $\varphi_{t,i}$ sampled from families of candidate spatial transformations $\Phi_t$ to obtain a variant $u_{t,i}$, which will be used with $u_{o,i}$ for TC learning.}  
    \label{fig:stc} 
    \vspace{-12pt}
\end{figure*}

Motivated by the observation in supervised learning that factoring 3D filters into separate spatial and temporal components yields significant gains~\cite{tran2018closer}, we propose to decouple the overall objective of unsupervised spatial-temporal feature learning into separate subtasks and provide regularizations to guide them to emphasize spatial and temporal
features, respectively (Fig.~\ref{fig:stc}).


\textbf{Spatial Contrast} is designed to focus on learning spatial
representations.  Neural networks are notorious for learning shortcuts
to ``cheat''~\cite{doersch2015unsupervised,timearrow}.  While previous
work tries to avoid all
cheating~\cite{doersch2015unsupervised,timearrow,lee2017unsupervised,kim2019self},
we intentionally make use of this property. In Spatial Contrast, the
network is provided with a shortcut by augmenting clips with only
spatial transformations.  This allows the model to ``cheat'' by using
only spatial features to capture intra-instance similarity and
inter-instance difference. In other words, the network will not bother
with temporal features because capturing
spatial similarity is already enough for solving the matching task.

In particular, as shown in Fig.~\ref{fig:stc}, given a batch of clips $X=[x_1,x_2,...,x_B]$, each
from a different video, we augment each clip $i$ with
augmentations $\varphi_{o,i}$ and $\varphi_{s,i}$ sampled from
families of candidate spatial transformations $\Phi_o$ and $\Phi_s$ 
to obtain corresponding variants $u_{o,i}= \varphi_{o,i}(x_i),$
$\varphi_{o,i}(\cdot)\sim\Phi_o$ and $u_{s,i} = \varphi_{s,i}(x_i),$
$\varphi_{s,i}(\cdot)\sim\Phi_s$, which are only spatially-augmented.


Our primary goal is to train an encoder so that the similarity between
feature embeddings of $u_{o,i}$ and $u_{s,j}$ is maximized when $i=j$,
and minimized otherwise.  Let $f(\cdot)$ denote the encoder. The
embeddings of $u_{o,i}$ and $u_{s,j}$ are $v_{o,i} = f(u_{o,i})$ and
$v_{s,j} = f(u_{s,j})$, where $v_{o,i}$ and $v_{s,j}$ are both vector
embeddings, e.g., obtained through global pooling.
%
%
%
%
%
By measuring similarity with cosine distance
$sim(v_{o,i},v_{s,j}) = (v_{o,i}\cdot
v_{s,j})/(|v_{o,i}|\cdot|v_{s,j}|)$, our goal can be achieved by
optimizing a contrastive loss called
InfoNCE~\cite{oord2018representation},
\begin{equation}
    L_{s} = -\sum_{i=1}^{B}\log\frac{\exp(sim(v_{o,i},v_{s,i})/\tau)}{\sum_{j=1}^{B} \exp(sim(v_{o,i},v_{s,j})/\tau)},
    \label{eq:6}
\end{equation}
\noindent
where $\tau$ is a temperature  controlling the concentration
of the feature embedding distribution~\cite{hinton2015distilling} (usually around $0.1$).
Eq.~\ref{eq:6} can be viewed as a cross entropy loss
between a pseudo prediction $S \in R^{B \times B}$ and a pseudo label
$\Bar{S} \in R^{B \times B}$.  $\Bar{S}$ is an identity matrix and $S$
is a similarity matrix where $S_{ij}$ measures the similarity of one
clip variant $u_{o,i}$ to another variant $u_{s,j}$ in the feature
space,
\begin{equation}
    S_{ij} = \frac{\exp(sim(v_{o,i},v_{s,j})/\tau)}{\sum_{m=1}^{B} \exp(sim(v_{o,i},v_{s,m})/\tau)}.
    \label{eq:7}
\end{equation}

In this spirit, the Spatial Contrast subtask is a self-supervised
multi-way classification problem in which we want to match one clip
variant to another when they come from the same video and distinguish
it from variants coming from different videos.


\textbf{Temporal Contrast} emphasizes learning temporal
representations. To do this, we need to prevent the network from
cheating through spatial feature similarity: we want variants whose
spatial context varies dramatically from the original clip, but whose
temporal context remains nearly the same and thus is essential for
capturing instance-level invariance. To do this, we add temporal
augmentations of random temporal cropping before applying spatial
transformations to produce another temporally-augmented variant. Note
that spatial augmentations are crucial here to further alter spatial
context in order to prevent cheating, and we show its importance
in Section~\ref{sec:ablation}.

Specifically, to produce a temporally-augmented variant
$u_{t,j}$ for each clip $x_j$, random temporal cropping
$\Gamma(\cdot)$ is first applied, followed by spatial augmentations
$\varphi_{t,j}$ sampled from another family of candidate spatial
transformations $\Phi_t$, after which we have $u_{t,j} =
\varphi_{t,j}(\Gamma(x_j)), ~\varphi_{t,j}(\cdot)\sim\Phi_t$.
%
%
%
%
%
Temporal Contrast is modeled between $u_{o,i}$ and $u_{t,j}$
using a similar technique as in Spatial Contrast. Then
we minimize,
\begin{equation}
    L_{t} = -\sum_{i=1}^{B}\log\frac{\exp(sim(v_{o,i},v_{t,i})/\tau)}{\sum_{j=1}^{B} \exp(sim(v_{o,i},v_{t,j})/\tau)}.
    \label{eq:9}
\end{equation}
As with Spatial Contrast learning, the
Temporal Contrast subtask is also a self-supervised multi-way
classification problem of matching clip variants of the same video.

\subsection{Hierarchical Contrast}
\label{sec:histc}
\begin{figure*}[t]
    \begin{center}
       \includegraphics[width=0.85\textwidth]{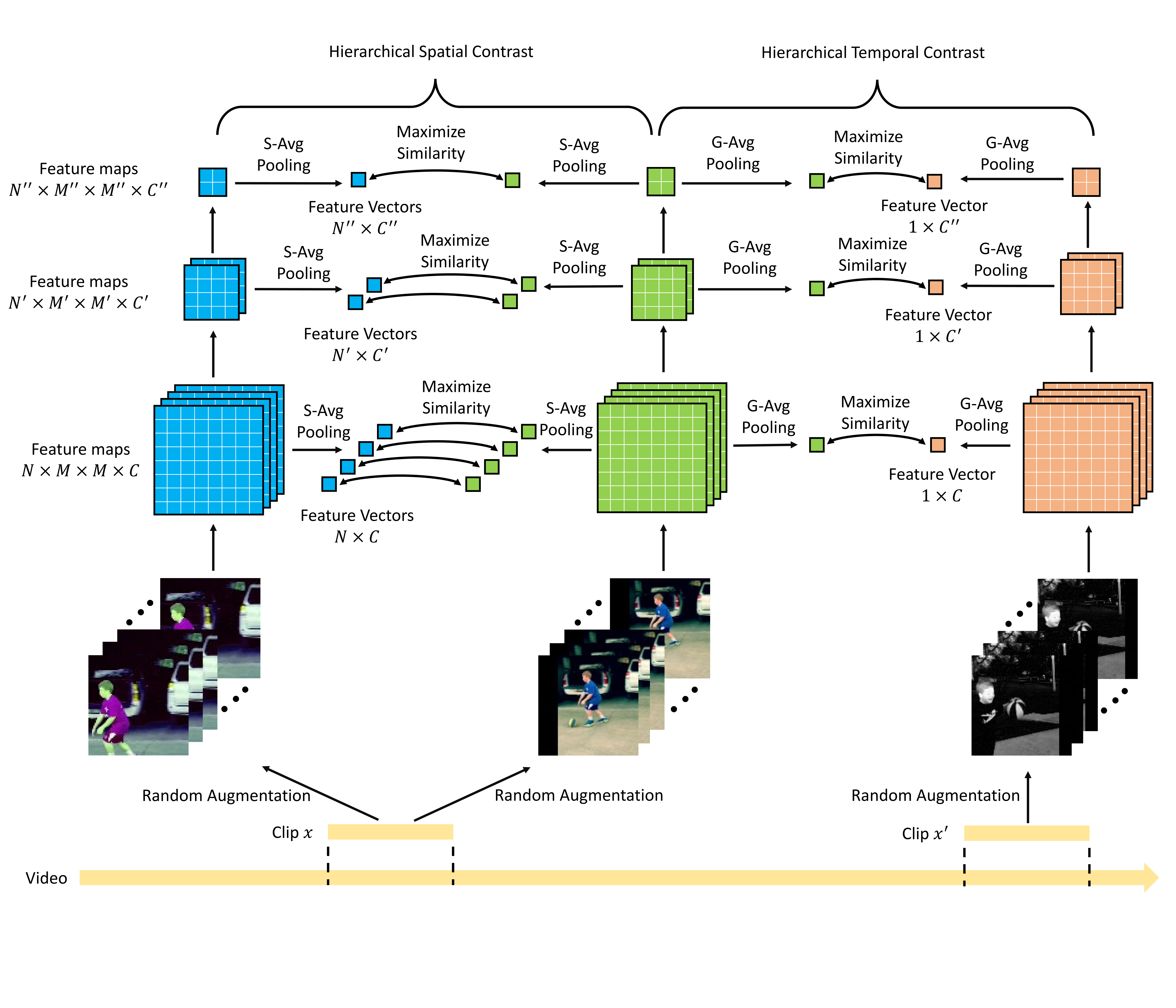}
    \end{center}
    \vspace{-12pt} 
    \caption{An example illustrating Hierarchical Contrast. Here we only show learning among variants from the same video and thus the network is maximizing similarity between all corresponding pairs. Otherwise if variants are from different videos, the network will minimize similarity.
    Feature maps at multiple scales are 2D global average pooled along the 
    spatial dimension (\textit{S-Avg Pooling}) in Hierarchical Spatial Contrast 
    learning, and 3D global average pooled along the temporal and spatial dimensions (\textit{G-Avg Pooling}) in Hierarchical Temporal Contrast learning. The channel dimension is omitted for clarity.}  
    \label{fig:histc} 
    \vspace{-12pt}
\end{figure*}

Inspired by the effectiveness of multi-scale features in supervised learning~\cite{zhou2018temporal, hussein2019timeception}, we introduce hierarchical learning to unsupervised learning by conducting Decoupled
Contrast learning hierarchically. As illustrated in
Fig.~\ref{fig:histc}, feature maps from different layers or blocks of
the encoder $f(\cdot)$ are collected and pooled to produce multi-scale
vector embeddings. Two pooling strategies are applied  to fully
leverage instance-level consistency among different layers: (1) 
Temporal Contrast uses 3D global average pooling along both
temporal and spatial dimensions, and (2) Spatial Contrast performs
2D global average pooling only along the spatial dimension. Therefore,
for each scale, we obtain one vector representation for each clip
variant in Hierarchical Temporal Contrast learning, but may get more
than one vector for each variant in Hierarchical
Spatial Contrast learning, depending on the length of input clip and
the temporal downsample factor of the encoder for a certain layer. Our
motivation is that because the timestamps in Spatial Contrast are
the same for two clip variants from the same video, the
spatial instance-level invariance should exist not only for the whole
clip, but also for corresponding sub-clips.

As shown in Fig.~\ref{fig:histc}, for Hierarchical Spatial Contrast learning at scale $k$, 
we apply 2D global average pooling and
obtain vector embeddings of $v^{k,n}_{o,i}$, $n=1,2,...,N$ for clip
variant $u_{o,i}$ and vector embeddings of $v^{k,n}_{s,j}$,
$n=1,2,...,N$ for variant $u_{s,j}$, where $N$ is the size
of the temporal dimension of the feature maps.  The loss 
at scale $k$ becomes,
\begin{equation}
    L^k_{s} = -\frac{1}{N}\sum_{n=1}^{N}\sum_{i=1}^{B}\log\frac{\exp(sim(v^{k,n}_{o,i},v^{k,n}_{s,i})/\tau)}{\sum_{j=1}^{B} \exp(sim(v^{k,n}_{o,i},v^{k,n}_{s,j})/\tau)},
    \label{eq:10}
\end{equation}
\noindent
while the loss for
Hierarchical Temporal Contrast remains
almost the same for scale $k$,
\begin{equation}
    L^k_{t} = -\sum_{i=1}^{B}\log\frac{\exp(sim(v^k_{o,i},v^k_{t,i})/\tau)}{\sum_{j=1}^{B} \exp(sim(v^k_{o,i},v^k_{t,j})/\tau)},
    \label{eq:11}
\end{equation}
where $v^k_{o,i}$ are $v^k_{t,i}$ are obtained from 3D global average pooling.

\subsection{Self-supervised Learning with HDC}

We consider 3 different backbones, C3D~\cite{c3d}, 3D-ResNet18~\cite{resnet,3dresnet} and R(2+1)D-10~\cite{tran2018closer}, as our
encoder $f(\cdot)$ for fair comparison with existing methods.  All
fully-connected layers are removed (last 3 layers of C3D, and last
layer of 3D-ResNet18 and R(2+1)D-10). Batch normalization layers~\cite{batchnorm} in
3D-ResNet18 and R(2+1)D-10 are replaced with instance normalization
layers~\cite{ulyanov2016instance} to prevent cheating via batch
statistics~\cite{henaff2019data,he2019momentum}; for C3D, an instance
normalization layer is inserted after each convolution layer. For
multi-scale learning, we consider feature maps from blocks
$3, 4, 5$ (last 3 blocks). Instead of directly performing the matching pretext task
with the feature vectors from pooling, we project each vector to a
lower-dimension space with linear projections,
following~\cite{wu2018unsupervised}. In particular, there are
projections $g^k_o(\cdot)$, $g^k_s(\cdot)$, and $g^k_t(\cdot)$ for
projecting $v^k_{o,i}$, $v^k_{s,i}$, and $v^k_{t,i}$, respectively, at
each scale $k$, each implemented as a single fully-connected layer
with linear activation projecting a vector into 128-d. 
(See~\cite{chen2020simple} for more experiments and discussion about
the importance of such linear projections.) The families of spatial
augmentations $\Phi_o$, $\Phi_s$, and $\Phi_t$ contain the same set of
transformations of random spatial cropping, scale jittering,
horizontal flipping, color jittering, and channel
replication~\cite{lee2017unsupervised}. We use $\tau=0.07$ for computing the contrastive loss.

We perform self-supervised training on
Kinetics-400~\cite{kay2017kinetics}, which has 400 action classes
and over 400 videos per class.  We resize frames,
preserving aspect ratio, so that the shorter
side is 128 pixels.  Each mini-batch contains 128 clips from 128 videos
and each clip consists of 16 randomly-cropped continuous frames of
shape $128 \times 128 \times 3$. After spatial and temporal
augmentations, three sets of clip variants are obtained, each of shape
$128 \times 16 \times 112 \times 112 \times 3$.  Before being fed to
the encoder, these clips are rescaled following~\cite{i3d} so
 pixel values are between -1 and 1.

Our model is implemented with Tensorflow~\cite{tensorflow} and
Keras~\cite{keras}. We use stochastic gradient descent with learning
rate $0.1$, momentum $0.9$, decay $0.0001$, and $L2$ regularizer
$5e^{-5}$.
HDC is trained as a whole towards minimizing a novel compound contrastive loss, namely HD-NCE,
\begin{equation}
    L = \sum_k (\alpha_k \cdot L^k_s + \beta_k \cdot L^k_t),
    \label{eq:12}
\end{equation}
where $L^k_s$ and $L^k_t$ are in Eqs.~\ref{eq:10}
and~\ref{eq:11}, and $k=3,4,5$ as we use features from blocks $3, 4, 5$ of
the encoder.

\xhdr{Significance of instance-wise
consistency.} $\alpha_k$ and $\beta_k$ represent
the significance of instance-wise
consistency in each layer, essentially weighting how much the corresponding subtask will contribute to our main goal. We simply tested (1)
$\alpha_3=\beta_3=\alpha_4=\beta_4=\alpha_5=\beta_5=1.0$, and  (2)
$\alpha_3=\beta_3=0.25$, $\alpha_4=\beta_4=0.5$ and
$\alpha_5=\beta_5=1.0$, and the second set worked better (see
Section~\ref{sec:ablation}).

\subsection{Difference from Recent Work}
\label{sec:diff}
Some work~\cite{yang2020hierarchical, yang2020vthcl} also explores hierarchical
contrastive learning in videos. However, while HDC matches variants obtained with carefully designed augmentations, ~\cite{yang2020vthcl} brings closer features from the slow and fast streams of a Slowfast network~\cite{feichtenhofer2019slowfast}, and ~\cite{yang2020hierarchical} predicts future motion patterns. Moreover, our work introduces spatial-temporal learning decoupling, which is another novel component.

Another similar method is CVRL~\cite{cvrl}, which is similar in spirit to the variant of our model trained with only Temporal Contrast incorporated in the last layer. Our HDC decouples the learning into separate subtasks emphasizing spatial and temporal features and further performs the learning hierarchically. As later experiments (Sec.~\ref{sec:ablation} and Tab.~\ref{tab:ablation}) show, HDC achieves significant improvement over the CVRL-alike variant (the second row in Tab.~\ref{tab:ablation}).
\section{Experiments}
We follow a common protocol~\cite{misra2016shuffle} to evaluate the
effectiveness of our HDC by using the learned representations as
initialization and fine-tuning on the downstream task of action
recognition on UCF101~\cite{soomro2012ucf101} and
HMDB51~\cite{kuehne2011hmdb}.
UCF101~\cite{soomro2012ucf101} consists of 13,320 videos and 101 classes 
of human action. It has three train/test splits with a split ratio of about 7:3. 
HMDB51~\cite{kuehne2011hmdb} is another widely-used action
recognition dataset containing 6,766 videos and 51
classes. It also has three splits with a similar split
ratio as UCF101. In our ablation studies, if not explicitly mentioned, we report Top-1 accuracy on only UCF101 split 1. 
When comparing our method with other state-of-the-art, results 
averaged on three splits of UCF101 and HMDB51 are reported.

In fine-tuning, we use the same network (C3D, 3D-ResNet18, or R(2+1)D-10) as we did
in self-supervised learning.  A dropout layer~\cite{srivastava2014dropout} of ratio 0.5 is added after global average pooling, followed by a single fully-connected layer and softmax activation for classification. The instance
normalization layers are kept as they are.  Blocks 1-5 are initialized with the learned weights by self-supervised
training on Kinetics-400. The last layer is randomly initialized.
During fine-tuning, we use stochastic gradient descent with learning
rate $0.01$, momentum $0.9$, decay $0.0001$, and $L2$ regularizer
$5e^{-5}$. Each mini-batch contains 32 clips, each
with 16 continuous frames randomly cropped to $128\times128\times3$.
Augmentations including random spatial cropping,  scale
jittering, and horizontal flipping are applied,
resulting in an input of shape $32\times16\times112\times112\times3$.

During testing, each video is divided into non-overlapping 16-frame
clips.  A center crop and four corner crops are taken for each
clip~\cite{wang2015towards}.  The class score for each video is
obtained by averaging over all crops and clips.


\subsection{Ablation Studies}
\label{sec:ablation}
We conduct ablation experiments to analyze our design choices.

\begin{table}
      \vspace{-12pt}
  {\footnotesize{\textsf{
\begin{center}
\scalebox{1.0}{
      \begin{tabular}{ccccc}
\toprule
        SC  & TC & $\alpha_k$ & $\beta_k$ & Top-1 Acc\\  
\midrule
        5 & - & 1 & - &  61.3\\ 
        - & 5 & - & 1 & 62.7 \\ 
        5 & 5 & 1 & 1 & 65.5 \\ 
        4, 5 & 4, 5 & 1, 1 & 1, 1 & 66.9 \\ 
        4, 5 & 4, 5 & 0.5, 1 & 0.5, 1 & 67.8 \\ 
        3, 4, 5 & 3, 4, 5 & 1, 1, 1 & 1, 1, 1 & 66.8 \\ 
\midrule
        \multicolumn{2}{c}{Training from scratch} & - & - & 43.5  \\ 
\midrule
        3, 4, 5 & 3, 4, 5 & 0.25, 0.5, 1 & 0.25, 0.5, 1 & \textbf{69.0} \\ 
        \bottomrule
      \end{tabular}}
\end{center}
  }}}
  \vspace{-12pt}
    \caption{Ablation study  of decoupled contrast and hierarchical contrast. We use 3D-ResNet18 as the backbone, pretrain on Kinetics-400 and report top-1 accuracy on UCF101 split 1.  SC and TC indicate the
    scale at which we perform Spatial Contrast (SC) or Temporal
    Contrast (TC) self-supervised learning -- i.e., the index of the
    block we take features from.  $\alpha_k$ and $\beta_k$
    are defined in Eq.~\ref{eq:12} (listed in  order
    of scale).}
    \vspace{-12pt}
      \label{tab:ablation}
\end{table}

\begin{figure*}[t]
    \begin{center}
       \includegraphics[width=1.0\textwidth]{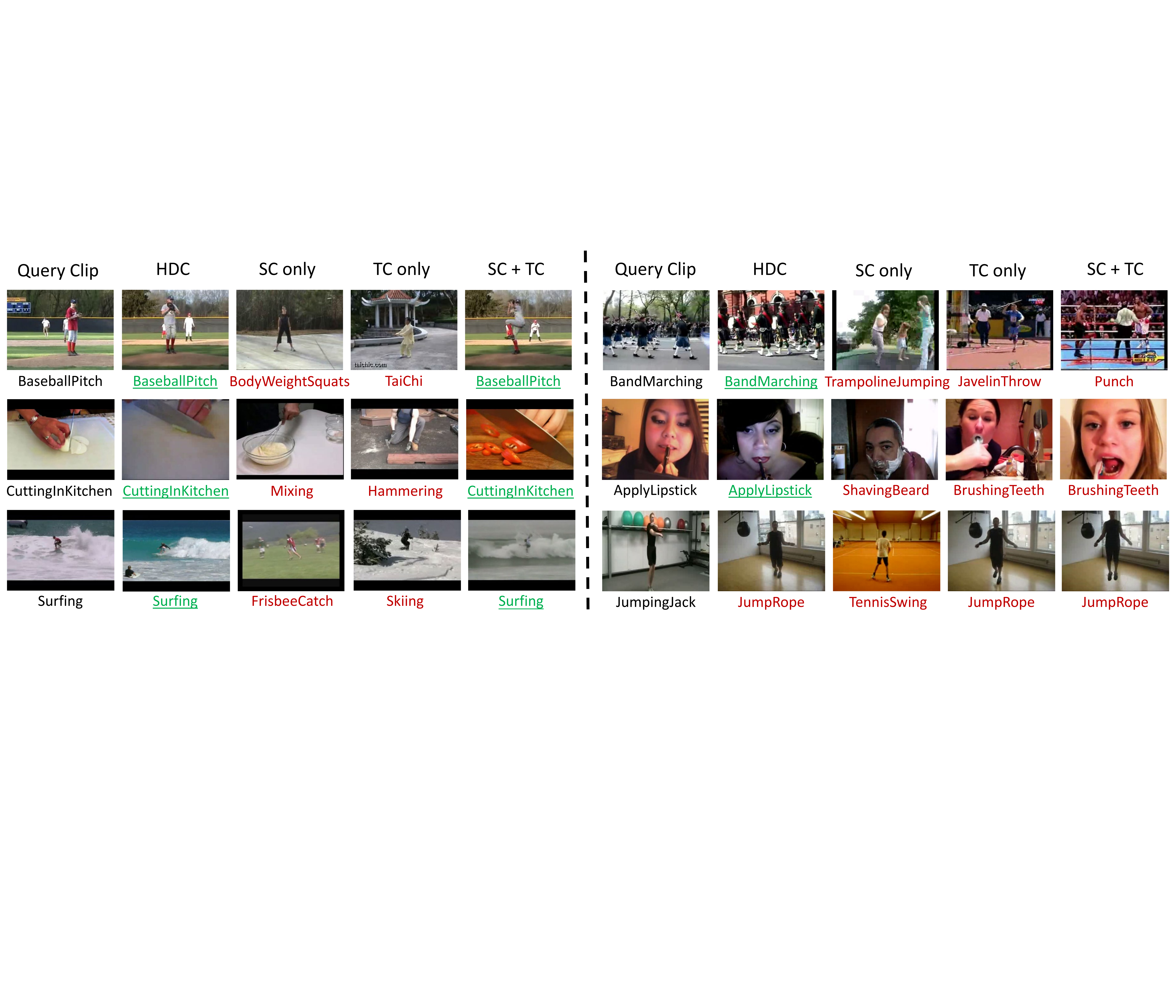}
    \end{center}
    \vspace{-18pt}
    \caption{Sample results of nearest neighbor retrieval on UCF101 split 1, showing the top-1 retrieved clip from our full model (\textit{HDC}), and Spatial Contrast  (\textit{SC}), Temporal
      Contrast  (\textit{TC}), and Joint Spatial-Temporal Contrast
       (\textit{SC + TC}) models. Below each clip is its 
      action category, with green underlined text indicating correct retrievals and red text indicating  incorrect. Our proposed HDC achieved better results because of the ability to capture both spatial and temporal features at multiple scales.}  
    \label{fig:retri} 
    \vspace{-12pt}
\end{figure*}

\xhdr{Decoupled Contrast.} We first evaluate the
effectiveness of decomposing the video representation learning task
into subtasks of Spatial and Temporal Contrast and performing joint
learning with them in Tab.~\ref{tab:ablation}.  The first three rows
present the results with only Spatial Contrast learning, with only
Temporal Contrast learning, and with both Spatial Contrast and
Temporal Contrast learning.  Hierarchical Contrast
learning is not involved and the learning is based on the features of
the last (5th) block. We see that, first, Temporal Contrast achieves slightly better performance than Spatial Contrast, perhaps because temporal semantics are more important in video learning. Then, by
incorporating both Spatial and Temporal Contrast, performance improves
significantly. This indicates that Spatial Contrast and Temporal
Contrast learn complementary features, and validates our hypothesis that
applying different augmentations as a way of regularization  guides
the network to learn different features. Furthermore, the nearest neighbor retrieval results in Fig.~\ref{fig:pre} and Fig.~\ref{fig:retri} show that Spatial Contrast and Temporal Contrast focus on spatial and temporal features respectively.

We note that Spatial Contrast in and of itself can be viewed as directly
migrating the basic contrastive learning model from the image to video
domain, while Temporal Contrast learning adapts to this new domain by
further applying temporal augmentations to create the sample
variant. Thus the results suggest that
traditional contrastive learning models should be further adapted to
learn a good embedding space for video, and we present
one possible solution.

\xhdr{Hierarchical Contrast.}
Comparing rows 3 with 4 or 6
of Tab.~\ref{tab:ablation}, we observe that
Hierarchical Contrast learning at more scales yields better performance.  This suggests that instance-level
invariance widely exists for mid-level and high-level features from
previous layers, and can be used  to capture multi-scale
semantics. However, rows 4 and 6 show that simply adding more scales does not consistently bring improvement. We argue that this is because instance-level
invariance may be weaker for mid-level features, and adding more scales while giving them the same weight of significance will distract the network and harm the learning. We discuss this 
below.

\xhdr{Significance of Instance-level Invariance at Different Scales.}
Zeiler et al.~\cite{zeiler2014visualizing} showed
that early layers of neural networks  learn low-level features
which change a lot due to augmentations. As we perform
Hierarchical Contrast learning at more scales, those mid-level
features may not share the same level of invariance against
augmentations as the last layer's features. We use  $\alpha_k$ and $\beta_k$ to model the significance of instance-level invariance at different
scales. We conducted experiments with different values of $\alpha_k$ and
$\beta_k$, and the results (row 4 vs row 5, row 6 vs the last row
in Tab.~\ref{tab:ablation}) show that smaller weights for lower
levels of the hierarchy yields better performance than assigning 1's. This
suggests that significance decreases at lower layers, which
is consistent with~\cite{zeiler2014visualizing}.
We did not exhaustively tune $\alpha_k$ and $\beta_k$, so
better performance is likely possible through tuning.

\begin{table}
      \vspace{-12pt}
  {\footnotesize{\textsf{
\begin{center}
\scalebox{0.85}{
      \begin{tabular}{ccccc}
\toprule
        color jittering  & channel replication & flipping & scale jittering & Top-1 Acc\\  
\midrule
        \xmark &        &        &        & 65.4\\ 
               & \xmark &        &        & 61.0 \\ 
        \xmark & \xmark &        &        & 49.4 \\ 
               &        & \xmark &        & 68.0 \\ 
               &        &        & \xmark & 64.7 \\ 
\midrule
        \multicolumn{4}{c}{Our full model (HDC)} & \textbf{69.0}  \\
        \bottomrule
      \end{tabular}}
\end{center}
  }}}
  \vspace{-12pt}
    \caption{Ablation study  of different spatial augmentations. We use our full  HDC model with 3D-ResNet18 as the backbone, pretrain on Kinetics-400, and show top-1 accuracy on UCF101 split 1. An \xmark~indicates the  transformation is not used during pretraining to generate augmented variants.}
    \vspace{-12pt}
      \label{tab:augab}
\end{table}

\xhdr{Spatial augmentation ablations.} Tab.~\ref{tab:augab} shows
ablation results of using different spatial augmentations. We find
that channel replication is crucial for the model to learn good
features, perhaps because it is a non-linear projection of RGB channels
which can effectively prevent the network from learning trivial
solutions based on color distribution~\cite{lee2017unsupervised}. As
another way to prevent such trivial solutions, color jittering
uses a linear function and thus is less effective. However, when
neither color jittering nor channel replication is used, the accuracy
drops greatly, indicating the network may suffer from the trivial
solution.

We note that, as shown in~\cite{chen2020simple}, there are other
spatial augmentations which can further improve the performance of
contrastive learning. However, the purpose of this paper is not to
exhaustively explore effects of different augmentations. By applying a
set of simple augmentations, we show the generalizability of our HDC.

\xhdr{Spatial augmentations in Temporal Contrast.} We verify the
importance of applying spatial augmentations after temporal random
cropping in Temporal Contrast. We use our full model of HDC with C3D
as the backbone, pretrain on Kinetics-400, and report average accuracy
on 3 splits of UCF101 and HMDB51. It achieves $72.3\%$ on UCF101 and
$39.3\%$ on HMDB51 when spatial augmentations are used in Temporal
Contrast versus only $68.9\%$ on UCF101 and $38.0\%$ on HMDB51
when spatial augmentations are not used. This supports our design
choice of having spatial augmentations in Temporal Contrast to obtain
variants whose spatial context varies as much as possible.

\setlength{\tabcolsep}{2pt}
\begin{table}
        \centering

      \vspace{-12pt}
  {\footnotesize{\textsf{
\begin{center}
\scalebox{0.9}{
      \begin{tabular}{lccccc}
\toprule
\multicolumn{2}{l}{Self-supervised Learning} & & \multicolumn{2}{c}{Accuracy ($\%$)} \\
\cmidrule{1-2} \cmidrule{4-5}
        Method  & Architecture & & HMDB51 & UCF101\\  
\midrule
        Random Initialization~\cite{misra2016shuffle} & CaffeNet && 13.3 & 38.6 \\
        Shuffle \& Learn~\cite{misra2016shuffle} & CaffeNet && 18.1 & 50.2 \\ 
        B{\"u}chler et al.~\cite{buchler2018improving} & CaffeNet && 25.0 & 58.6 \\
\midrule
        Random Initialization~\cite{lee2017unsupervised} & VGG-M-2048 && 18.3 & 51.1 \\
        OPN~\cite{lee2017unsupervised} & VGG-M-2048 && 23.8 & 59.8 \\
\midrule
        Random Initialization~\cite{xu2019self} & R3D-18 && 21.5 & 54.4 \\
        Clip Order~\cite{xu2019self} & R3D-18 && 29.5 & 64.9 &\\
        VCP~\cite{luo2020video} & R3D-18 && 31.5 &  66.0 \\
\midrule
        Random Initialization~\cite{cyclecontrast} & R3D-18+1 && 19.4 & 44.7 \\
        CCL~\cite{cyclecontrast} & R3D-18+1 && 37.8 & 69.4 \\
\midrule
        Random Initialization~\cite{han2019video} & 2D3D-ResNet18 && 17.1 & 46.5 \\
        DPC~\cite{han2019video} & 2D3D-ResNet18 && 34.5 & 68.2 \\
\midrule
        Random Initialization~\cite{benaim2020speednet} & I3D && \textbf{29.6} & 47.9 \\
        SpeedNet~\cite{benaim2020speednet} & I3D && \textbf{43.7} & 66.7 \\
\midrule
        Random Initialization~\cite{3dresnet} & 3D-ResNet18 && 17.1 (19.1) & 42.4 (43.8) \\
        3D-RotNet~\cite{rotnet} & 3D-ResNet18 && 33.7 & 62.9 \\
        3D-ST-Puzzle~\cite{kim2019self} & 3D-ResNet18 && 33.7 & 65.8 \\
\midrule
        \textbf{Our method (HDC\textdagger)} & \textbf{3D-ResNet18} && \textbf{38.4} & \textbf{69.8} \\
        \textbf{Our method (HDC)} & \textbf{3D-ResNet18} && \textbf{38.1} & \textbf{68.5} \\
\midrule
        Random Initialization~\cite{xu2019self} & C3D && 23.2 (22.5) & 61.6 (51.4) \\
        Motion \& Appearance~\cite{wang2019self} & C3D && 33.4 & 61.2 \\ 
        Clip Order~\cite{xu2019self} & C3D && 28.4 & 65.6 \\
        VCP~\cite{luo2020video} & C3D && 32.5 & 68.5 \\
        Cho et al.~\cite{cho2020self} & C3D && 34.3 & 70.4 \\
\midrule
        \textbf{Our method (HDC\textdagger)} & \textbf{C3D} && \textbf{39.5} & \textbf{73.1} \\
         \textbf{Our method (HDC)} & \textbf{C3D} && \textbf{39.3} & \textbf{72.3} \\
\midrule
        Random Initialization~\cite{xu2019self} & R(2+1)D-10 && 22.0 & 56.2 \\ 
        Clip Order~\cite{xu2019self} & R(2+1)D-10 && 30.9 & 72.4 \\
        VCP~\cite{luo2020video} & R(2+1)D-10 && 32.2 & 66.3 \\
        Cho et al.~\cite{cho2020self} & R(2+1)D-10 && 36.8 & 74.8 \\
        PacePrediction~\cite{pacepredict} & R(2+1)D-10 && 36.6 & \textbf{77.1} \\
\midrule
        \textbf{Our method (HDC\textdagger)} & \textbf{R(2+1)D-10} && \textbf{40.0} & \textbf{76.8} \\
        \textbf{Our method (HDC)} & \textbf{R(2+1)D-10} && \textbf{39.8} & \textbf{76.2} \\
        \bottomrule
      \end{tabular}}
\end{center}
  }}}
  \vspace{-6pt}
        \caption{Top-1 accuracy averaged on 3 splits of
        UCF101 and HMDB51. Parentheses show accuracy obtained with
        our own implementation. HDC with~\textdagger~is pretrained by decreasing learning rate more slowly.}
      \label{tab:compa}
      \vspace{-12pt}
\end{table}
\setlength{\tabcolsep}{4pt}
\begin{table}
    \centering

      \vspace{-12pt}
  {\footnotesize{\textsf{
\begin{center}
\scalebox{1.0}{
      \begin{tabular}{lccccc}
\toprule
        Methods  & Top1 & Top5 & Top10 & Top20 & Top50\\  
\midrule
        OPN~\cite{lee2017unsupervised} & 19.9 & 28.7 & 34.0 & 40.6 & 51.6 \\
        B{\"u}chler et al.~\cite{buchler2018improving} & 25.7 & 36.2 & 42.2 & 49.2 & 59.5 \\ 
\midrule
        Random Initialized C3D & 16.7 & 27.5 & 33.7 & 41.4 & 53.0 \\
        Clip Order (C3D)~\cite{xu2019self} & 12.5 & 29.0 & 39.0 & 50.6 & 66.9 \\
        VCP (C3D)~\cite{luo2020video} & 17.3 & 31.5 & 42.0 & 52.6 & 67.7 \\
\midrule
        CCL (R3D-18+1)~\cite{cyclecontrast} & 22.0 & 39.1 & 44.6 & 56.3 & 70.8 \\
\midrule
        \textbf{Our HDC (C3D)} & \textbf{33.9} & \textbf{49.6} & \textbf{55.7} & \textbf{61.6} & \textbf{69.9} \\ 
        \bottomrule
      \end{tabular}}
\end{center}
  }}}
    \vspace{-12pt}
        \caption{Nearest neighbor retrieval results on UCF101.}
      \label{tab:nnrucf}
\end{table}
\begin{table}
      \vspace{-12pt}
  {\footnotesize{\textsf{
\begin{center}
\scalebox{1.0}{
      \begin{tabular}{lccccc}
\toprule
        Methods  & Top1 & Top5 & Top10 & Top20 & Top50\\  
\midrule
        Random Initialized C3D & 7.4 & 20.5 & 31.9 & 44.5 & 66.3 \\
        Clip Order (C3D)~\cite{xu2019self} & 7.4 & 22.6 & 34.4 & 48.5 & 70.1 \\
        VCP (C3D)~\cite{luo2020video} & 7.8 & 23.8 & 35.3 & \textbf{49.3} & \textbf{71.6} \\
\midrule
        \textbf{Our HDC (C3D)} & \textbf{14.6} & \textbf{28.8} & \textbf{36.1} & \textbf{44.8} & \textbf{57.9} \\ 
        \bottomrule
      \end{tabular}}
\end{center}
  }}}
  \vspace{-12pt}
        \caption{Nearest neighbor retrieval results on HMDB51.}
        \label{tab:nnrhmdb}
    \vspace{-12pt}
    
\end{table}

\subsection{Comparison with the State-of-the-art}
For fairness, we  compare with methods under similar settings. There are other methods that achieve great success. But they adopt much more advanced backbones~\cite{cvrl, yao2020seco, diba2019dynamonet}, require extra preprocessing~\cite{timearrow, Han20, han2020self}, or needs additional information to prepare the input~\cite{korbar2018cooperative,alwassel2019self, sun2019learning, miech2020end}. 

We report top-1 accuracy averaged over 3 splits of UCF101 and HMDB51
in Tab.~\ref{tab:compa}. HDC achieves better or comparable performance on both datasets. We note that: 1) PacePrediction~\cite{pacepredict} shows slightly better results on UCF101, perhaps because its best performance is achieved by further adapting contrastive laerning in addition to pace prediction; 2) SpeedNet~\cite{benaim2020speednet} benefits from the backbone ($29.6\%$ when trained from scratch) and thus has better performance on HMDB51.

By comparing HDC variants using
different backbones, we observe that a more advanced backbone always
leads to better performance, suggesting that HDC will be able to further benefit from future advances in network architectures. We also find that decreasing the learning rate more slowly is beneficial, perhaps because it allows better optimization.



\subsection{Nearest Neighbor Retrieval}
\label{sec:retrieval}
We follow~\cite{xu2019self} and perform nearest neighbor retrieval experiments. 
As shown in Tables~\ref{tab:nnrucf} and~\ref{tab:nnrhmdb}, our
method significantly outperforms other methods on both UCF101 and
HMDB51. This implies that we  learn better
features, and explains the good performance on downstream tasks.

As shown in Fig.~\ref{fig:pre} and Fig.~\ref{fig:retri}, qualitative
results further support our idea of manipulating augmentations to
guide the network to learn different features and the benefit of
hierarchical learning.  For example, in the first row of the right
column in Fig.~\ref{fig:retri}, HDC succeeds in retrieving a clip of
the correct action while the other variants fail: SC focuses more on
spatial information of people outdoors, TC pays more attention to the
actions of people walking, and ST+TC fails perhaps because it does not
learn features at different hierarchies, although it successfully
captured both spatial and temporal information of multiple moving
people.

\section{Conclusion}

We considered the problem of unsupervised video representation learning, and introduced 
Hierarchically Decoupled Spatial-Temporal Contrast (HDC). By decomposing the target into subtasks emphasizing different features and performing learning in a hierarchical manner, HDC is able to capture both rich spatial and temporal semantics at multiple scales. Extensive experiments of action recognition and nearest neighbor retrieval on UCF101 and HMDB51 using 3 different backbones suggest the potential of manipulating augmentations as regularization and demonstrate the state-of-the-art performance of HDC. 


{\small
\bibliographystyle{ieee_fullname}
\bibliography{mybib}
}

\end{document}